\documentclass{article}

\usepackage[nonatbib,final]{nips_2018}

\usepackage[utf8]{inputenc} 
\usepackage[T1]{fontenc}    
\usepackage{hyperref}       
\usepackage{url}            
\usepackage{booktabs}       
\usepackage{amsfonts}       
\usepackage{nicefrac}       
\usepackage{microtype}      
\usepackage{color}
\usepackage{graphicx}
\usepackage{amssymb}
\usepackage{amsmath}

\newcommand{\fig}[1]{Fig.~\ref{fig:#1}}

\newcommand{\secc}[1]{Section~\ref{sec:#1}}
\newcommand{\etal}{\textit{et al}.\:}

\title{Learning Goal Embeddings via Self-Play for Hierarchical Reinforcement Learning}

\author{
  Sainbayar Sukhbaatar,\hspace{3mm} Emily Denton \\
  Department of Computer Science\\
  New York University\\
  \texttt{\{sainbar,denton\}@cs.nyu.edu} \\
  \And
  Arthur Szlam,\hspace{3mm} Rob Fergus \\
  Facebook AI Research\\
  New York \\
  \texttt{\{aszlam,robfergus\}@fb.com}
}

\begin{document}

\maketitle

\begin{abstract}
In hierarchical reinforcement learning a major challenge is determining appropriate low-level policies. We propose an unsupervised learning scheme, based on asymmetric self-play from Sukhbaatar \etal \cite{Sukhbaatar18}, that automatically learns a good representation of sub-goals in the environment and a low-level policy that can execute them. A high-level policy can then direct the lower one by generating a sequence of continuous sub-goal vectors. We evaluate our model using Mazebase and Mujoco environments, including the challenging AntGather task. Visualizations of the sub-goal embeddings reveal a logical decomposition of tasks within the environment. Quantitatively, our approach obtains compelling performance gains over non-hierarchical approaches.
\end{abstract}

\section{Introduction}

Complex real-world tasks may require the agent to take thousands or millions of individual actions. However, tasks are usually a composition of a much smaller number of sub-tasks, reused across different objectives. This naturally suggests a hierarchical structure for the agent's controller, in which individual actions are performed by a low-level policy, while a higher-level one selects appropriate sub-goals. Correspondingly, hierarchical reinforcement learning (HRL) is a subject of much interest. Automatic learning of hierarchical policies poses many challenges, e.g. ensuring an appropriate division of labor between the levels without explicit supervision; preventing degeneracies in the low-level policies and deciding/designing how the levels should communicate with one another.

In this work we explore a form of HRL in which the high-level policy directs the low-level policy via a continuous sub-goal vector. With this approach, the challenge is to learn an appropriate representation space for the goals. Specifically, (i) the space should be general enough to cover the full range of sub-tasks within the environment; (ii) task-irrelevant details of the environment should be abstracted away, while retaining enough information to discriminate between different sub-goals and (iii) it should encode sub-goals achievable by the low-level policy (i.e. at the correct level of difficulty). 

Our approach uses unsupervised asymmetric self-play
 \cite{Sukhbaatar18} as a pre-training phase for the low-level policy, prior to training the hierarchical model. 
In self-play, the agent devises tasks for itself via the goal embedding and then attempts to solve them. Since each task involves a physical change in the environment, the goal embedding learned via self-play captures aspects of the environment that are  controllable, i.e. can be changed by the agent, and ignores parts that it cannot alter (e.g.~static background or purely random elements). This is helpful for the high-level policy since user-specified tasks typically involve manipulations of the environment.
Furthermore, the adversarial reward structure forces the agent to constantly come up with new tasks, thus ensuring a diverse goal representation. Imposing a time limit on each task limits their complexity and so ensures good coverage of goals of a given difficulty. 

A key aspect of our approach is the parameterization of the low-level policy, which takes as input both the current state and a goal vector. The latter is an encoding of the target state, learned during self-play to guide the agent to complete self-imposed tasks. This provides a natural mechanism for the higher level controller to specify sub-goals that make up complex tasks. The higher policy is trained using sparse task reward as supervision and we show experimentally that it is able to learn tasks that are difficult for an agent trained at the level of atomic actions.

\subsection{Related Work}



Options \cite{Sutton1999BetweenMA}, a formalization of temporal abstraction in an MDP, have become a popular framing of hierarchical RL.  While earlier works used pre-specified option policies, there has been recent success in discovering options.  For example, Bacon \etal \cite{Bacon17} extends the policy gradient theorem to the setting of options, and shows with the appropriate entropy regularization (to avoid collapsing to trivial one-option policies) and termination regularization (to avoid collapse onto full control of the low level actor by the high level controller), useful discrete options could be learned.  

Many recent works have considered options discovery via parameterized modules operating at different timescales, where an ``actor'' operates at a finer timescale than a ``manager'', which outputs a goal or target for the actor.  For example, Vezhnevets \etal \cite{Vezhnevets17} trains the actor and manager together end-to-end via reward from the environment.    

A line of work \cite{DBLP:conf/nips/MohamedR15, GregorRW16, Florensa17a,hausman18, Eysenbach18} takes this approach in the context of intrinsic motivation.
In Mohamed \etal \cite{DBLP:conf/nips/MohamedR15} a variational inference approach is used to make exploration via empowerment \cite{empowerment} tractable.   Continuing in this path \cite{GregorRW16, Florensa17a,hausman18, Eysenbach18}  use an actor parameterized by state and a latent vector in such a way that the latent vector is predictable from a final state or a sequence of states the actor visits, but otherwise, the actions have high entropy.   After pre-training in this way, a ``manager'' can learn to issue commands via the latent vector.  In Haarnoja \etal \cite{Haarnoja18}, a similar construction is used to train an agent end to end.   Our work also uses this construction,  but the unsupervised pre-training of the actor is done via asymmetric self-play as in \cite{Sukhbaatar18}.

There is a large literature on goal discovery and intrinsic motivation, both independent of RL \cite{Schmidhuber-ijcnn-91,OudeyerK09} and framed in terms of RL \cite{intrinsically_motivated_RL}.
Recently, Pete \etal \cite{pere18} used a construction where the goal space is learned first by using an auto-encoder on states from the environment, and then using a goal discovery algorithm on top of the learned representation.   In this work, we use an intrinsic motivation approach to learn both a low level actor and the representation of the state space.  In future work, we intend to do as in \cite{pere18} and consider goal discovery at the level of the manager as well. 






\begin{figure}[h!]
	    \centerline{%
		    \includegraphics[width=\linewidth]{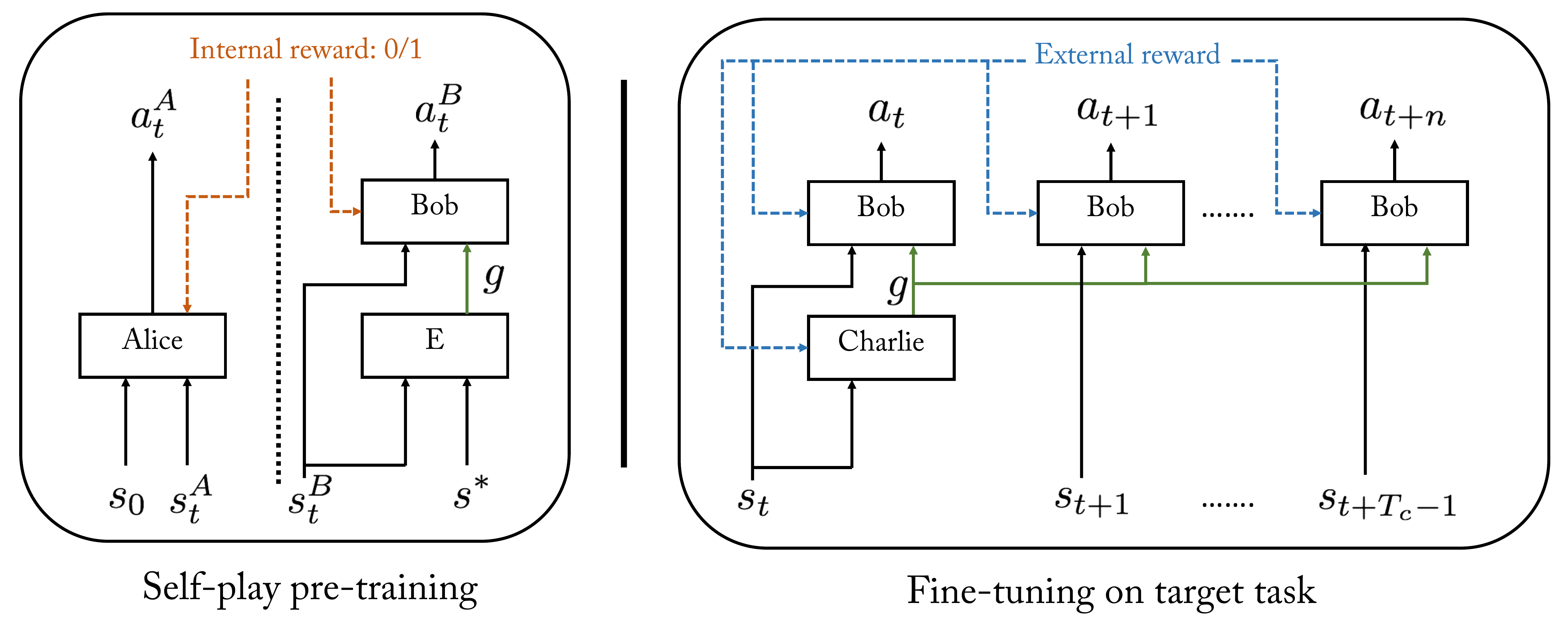}
		 }
		\caption{The Hierarchical Self-Play model in pre-training (left) and fine-tuning (right) phases.}
		\label{fig:model}
\end{figure}

\section{Approach}

We consider single-agent reinforcement learning in a fully-observable MDP. Let $s \in \mathbb{R}^s$ be the state, $a$ be an action. The agent has a hierarchical controller that consists of a high-level policy $\pi_C$, known as "Charlie", that directs a low-level actor "Bob" $\pi_B$. Both $\pi_B$ and $\pi_C$ are modeled with deep neural nets. On a given target task, both the high and low-level policies will be learned using external reward after an unsupervised pre-training phase.  

The pre-training phase consists of Bob exploring and building skills.   Bob is paired with another controller $\pi_A$ "Alice". During the pre-training phase, Bob learns a continuous goal embedding $g$ which will be used by Charlie on the target task to communicate local sub-goals to Bob. \fig{model} summarizes the approach. 

\subsection{Self-play Pre-training}
In this phase, Alice and Bob take turns in controlling the agent. An internal reward is structured to induce a competition between them that results in the agent exploring the environment. In so doing, Bob gains an understanding of how the environment operates, i.e.~how to transition from one state to one nearby. The approach is a modified form of asymmetric self-play introduced in Sukhbaatar \cite{Sukhbaatar18}. 

Self-play commences with $\pi_A$ and $\pi_B$ randomly initialized and the agent in some initial state $s_0$. First, Alice takes $T_A$ steps in the environment using policy $\pi_A$. 
\[ a_t^A = \pi_A(s_t^A, s_0) \]
Let $s^* = s_{T_A}^A$ be Alice's final state, which becomes Bob's goal. Next, we reset the environment back to $s_0$ and Bob takes control, taking actions according to his policy $\pi_B$. 
\[ a_t^B = \pi_B(s_t^B, s^*) \]
Bob is deemed to have succeeded if any time step his state is close to $s^*$ under distance function $D$ (which depends on the environment; see \secc{experiments}): 
\[ D(s_t^B, s^*) \le \epsilon \] 
If after $T_B$ steps this criterion has not been met, Bob has failed.  Bob's reward $R_B$ is 1 if he succeeds, 0 otherwise. Alice's reward is $R_A=1 - R_B$. This reward is used to update $\pi_A$ and $\pi_B$ using Reinforce \cite{Williams92simplestatistical}, although any other policy gradient or Q-learning based algorithm could potentially be used. The reward structure causes Alice to seek states $s^*$ that Bob has difficulty reaching. But since we choose $T_A \sim T_B$ and both have the same abilities (in terms of capacity \& actions), Bob will quickly master a task set by Alice, forcing her to find new, unexplored ones to challenge Bob. Through many episodes of self play, Alice and Bob are thus able to explore the environment effectively. 

In self-play, Bob is trained only on tasks that are achievable under $T_A$ steps. This becomes an issue when episodes always start at the same state, because Bob never observes states that are far from the initial state $s_0$. A simple solution is to perform several self-play games within a single episode. An episode starts with a standard self-play game where Bob tries to reach the same state as Alice. If Bob succeeds, then the episode continues with another self-play game. However, for the next game, both Alice and Bob start from the last states they reach in the previous game. This allows them to explore far away states. An episode ends either when Bob fails or after a fixed number of games. Rewards are discounted by a hyperparameter $\lambda$ between games, while the reward discount is 1 within each game. 

Since, for every proposed task, we have the ground truth actions from Alice for achieving that task, we can train Bob's policy to imitate them. Thus, the final form of Bob's loss function is
\[L_B = \mathbb{E}_{a_t^B \sim \pi_B} [-R_B] + \alpha \mathbb{E}_{a_t^A \sim \pi_A} [-\log(\pi_B(a_t^A| s_t^A))],\]
where the hyper-parameter $\alpha$ is for balancing the two loss terms.

For Alice, we add an entropy regularization on her policy  to encourage her to propose diverse tasks. Her loss function is
\[L_A = \mathbb{E}_{a_t^A \sim \pi_A} [-R_A - \beta \mathrm{H}(\pi_A(s_t^A))].\]
Here $\mathrm{H}$ is the entropy function, and $\beta$ is the coefficient of the entropy regularization.

This scheme differs from \cite{Sukhbaatar18} in several ways:
\paragraph{(i)} The number of steps taken by Alice and Bob are fixed to $T_A$ and $T_B$ respectively, versus being dynamic as in \cite{Sukhbaatar18}. This constrains the scope of work done by the low-level policy to be manageable and also reduces the amount of exploration needed. We choose $T_B$ to be slightly larger than $T_A$ so Bob has a chance of success even with few mistakes. 
\paragraph{(ii)} The episodes are broken into multiple shorter segments with the environment reset to the beginning of the segment instead of the beginning of the episode.  This allows for more exploration while keeping Bob's policy manageable for Charlie.
\paragraph{(iii)} Keeping with the other changes, instead of having a reward for Bob based on time, we adopt a simplified 0/1 reward. 
\paragraph{(iv)} The structure of Bob, detailed below.

\subsection{Bob's Architecture}
Bob's policy has two components. The first is a goal encoder $E$ that maps the current and goal states to a low-dimensional goal embedding $g_t \in \mathbb{R}^K$:
\[ g_t = E(s^*, s_t^B) .\]
The low dimension of the space (i) acts as a bottleneck, forcing Bob to compactly represent the goal and (ii) makes Charlie's job easier, as he will be generating $g_t$'s to control Bob on the target task.

We consider two forms of the goal encoder $E$:
\begin{enumerate}
\item Compute the {\bf difference} between current and target states: $E(s^*, s_t^B) = \phi(s^*) - \phi(s_t^B)$, 
where $\phi$ is a state embedding function.
\item An {\bf absolute} representation, which just considers $s^*$: $E(s^*, s_t^B) = \phi(s^*) $. 
\end{enumerate}

Bob's second component is a policy conditioned on a goal embedding
\[a_t^B = \pi_B' (s_t^B, g_t).\]
Putting the two stages together, we have 
\[a_t^B = \pi_B (s_t^B, s^*) = \pi_B' (s_t^B, E(s^*, s_t^B)).\]

\subsection{Training Charlie}
After self-play pre-training, Bob is used as a low-level policy for solving the user-specified target task. A high-level policy, Charlie, is introduced that outputs a goal vector $g_t$ which controls Bob:
\[g_t = \pi_C(s_t).\]
This vector is then fed to Bob's goal policy $\pi_B'$, which converts it to an action on the environment
\[a_{t+i} = \pi_B'(s_{t+i}, g_t) .\]
Here, Bob will take multiple actions with the same goal for $i=[0, \cdots, T_C-1]$, thus Charlie's next action occurs at $t+T_c$. 
External reward from the task is used to train Charlie, as well as fine-tuning Bob's policy $\pi'_B$.
We choose $T_C$ to be equal to $T_A$ because Bob is trained on goals achievable in $T_A$ steps.

\subsection{Parameterization of the Components}
We use a multi-layer perceptron (MLP) with tanh non-linearity for parameterizing all the components of our model. 
Both $\pi_A$ and $\pi_C$ are two-layer MLPs. For $\phi$, we also use a two-layer MLP, but without non-linearity at the last layer to avoid putting bounds on the goal embedding. Lastly, we use a three-layer MLP for Bob's policy $\pi_B'$ where the second hidden layer is given by 
\[h_2 = \sigma(W_2\sigma(W_1 s_t) + W_g g_t).\]
$\sigma$ is a tanh non-linearity, and bias terms are omitted for brevity. All the hidden layers have 64 units.

All the policy networks have $M+1$ output heads, where $M$ is the dimension of the action space. For Charlie, $M$ is equal to the goal embedding dimension,  $K$. Besides $M$ action heads, each policy network also outputs a baseline value. For a discrete action space, each head is a linear layer followed by softmax. For continuous actions, each action head outputs $\mu$ and $\log(\sigma)$ values, and an action is sampled from $\mathcal{N}(\mu, \sigma)$.

\section{Experiments}
\label{sec:experiments}

We test our hierarchical self-play (HSP) model on two different environments. First, it is applied to a task procedurally generated in a grid-world environment, Mazebase \cite{mazebase}. The second environment is a control of an ant-like robot in Mujoco physics simulation \cite{mujoco}, where it has to collect randomly placed objects.

In all experiments, we set Alice's entropy regularization coefficient $\beta$ to $0.01$, and Bob's imitation coefficient $\alpha$ to $0.03$. For training all policies, we use the REINFORCE \cite{Williams92simplestatistical} algorithm with a learned baseline. However, our model can be trained by more sample efficient policy gradient algorithms such as TRPO and PPO. For optimization, we use RMSProp \cite{rmsprop} with learning rate 0.001 and $\alpha=0.97, \epsilon=1e-6$. 
We run each experiment 5 times with different random seeds, and report their mean and standard deviation. In the training plots, the pre-training steps are not included as they are unsupervised, and the same pre-trained model potentially can be used for different tasks. The code is available at \url{https://cims.nyu.edu/~sainbar/hsp}.

\subsection{MazeBase}
\begin{figure}[b!]
	    \centerline{%
		    \includegraphics[width=0.33\linewidth]{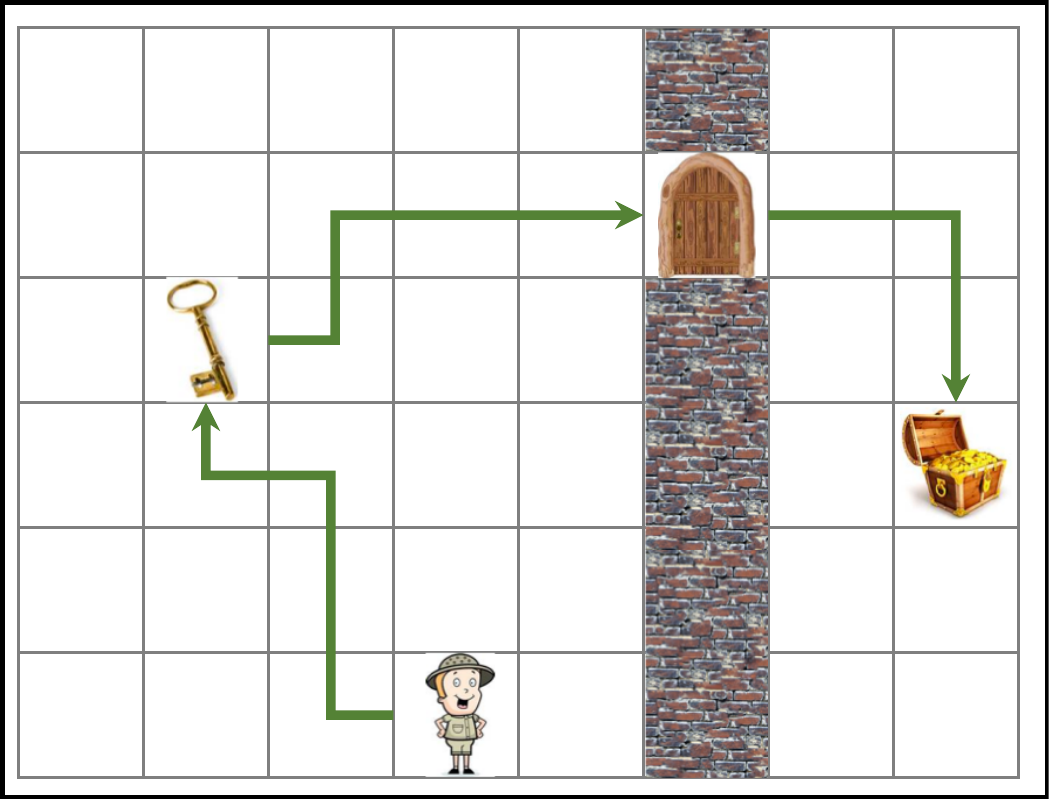} \hspace{5mm}
\includegraphics[width=0.5\linewidth]{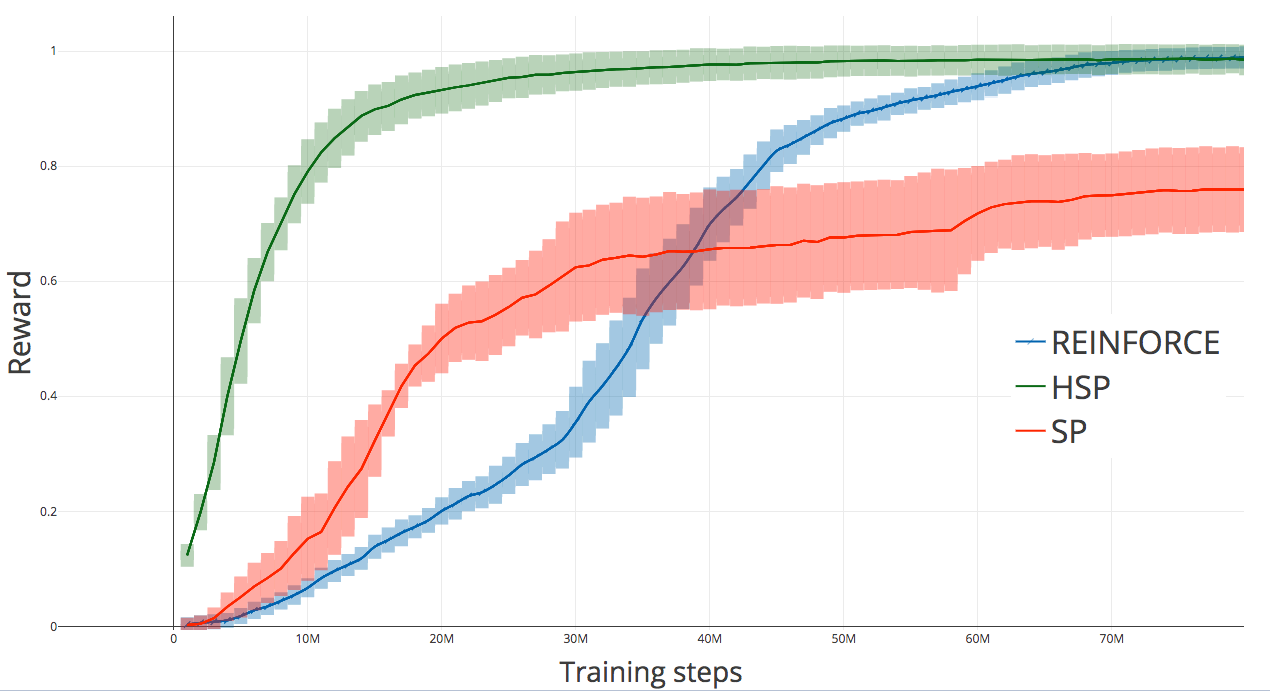}
		 }
		\caption{\textbf{Left: } In the KeyDoor task, the agent needs to unlock the door with the key and go to the goal. \textbf{Right:} Comparison of our approach HSP with other baselines: REINFORCE with no self-play (blue) and self-play from \cite{Sukhbaatar18} which has no hierarchy (red).}
		\label{fig:keydoor}
\end{figure}

We test our model in the Mazebase \cite{mazebase} environment on the ``Key-Door'' task, where the grid is divided into two rooms by a wall as shown in \fig{keydoor}(left). The objective is to reach the treasure goal, but the agent first needs to pick up the key and then open the door. 

The task is not trivial because the object locations are randomized for every episode. In addition, it has sparse reward ($+1$ for success and $0$ otherwise), making it even more challenging.

The observation is a binary vector of size $\text{MapWidth}\times \text{MapHeight}\times \text{VocabularySize}$. The vocabulary consists of words necessary for describing objects: ``agent'', ``door'', ``block'', ``key'' and ``goal''. The possible actions are four one-step movement actions, pick action, and stop action. To pick an item, the agent has to be on top of it and perform the pick action. Episodes terminate after $40$ steps. 

During self-play, we use the $\ell_2$-norm on the state as distance metric $D$ and set $\epsilon=0$. This means Bob needs to exactly match everything in the grid to Alice's final state. We use a goal embedding of $K=3$ dimensions. 

Alice always takes $T_A=5$ steps, and Bob has $T_B=7$ steps to reach the goal state.
However, we allow up to 4 self-play games per episode with discount of $\lambda=0.7$ between games. This effectively allows Alice and Bob to take up to $4 T_A=20$ and $4T_B=28$ steps respectively, which is usually enough for picking up the key and entering the other room. We use the absolute version of the goal embedding function $E$.
When training Charlie, we set $T_C=5$.

\begin{figure}[t!]
	    \centerline{%
\includegraphics[width=0.4\linewidth]{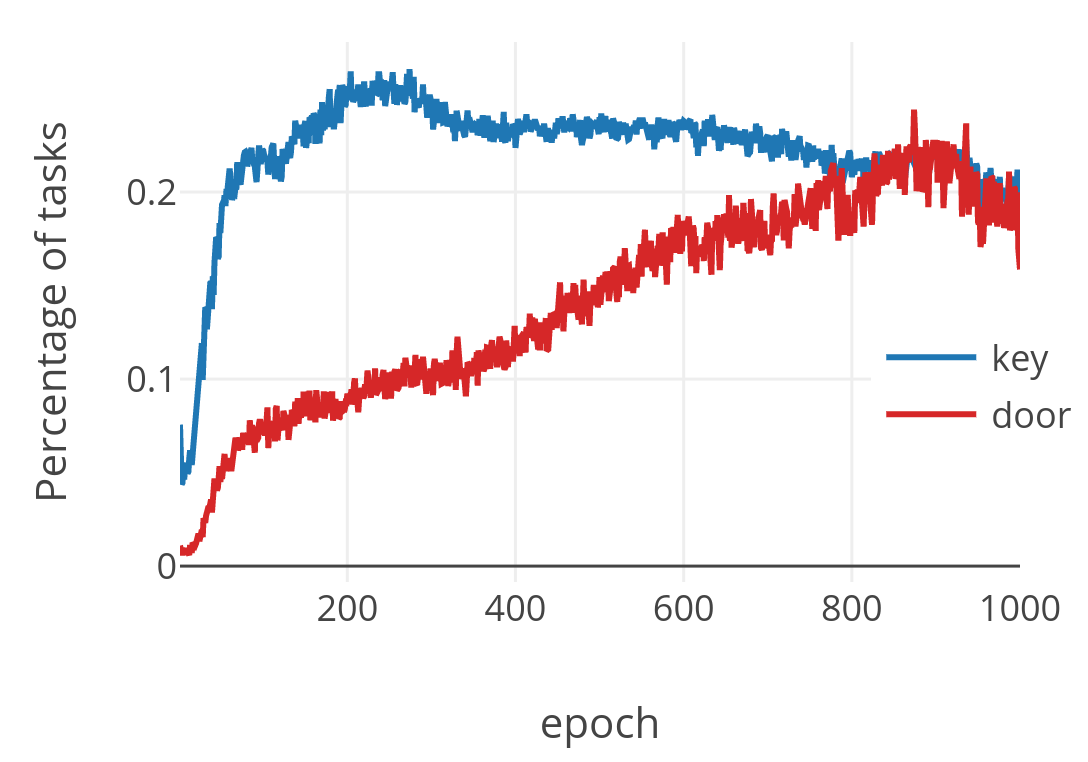} \hspace{5mm}
\includegraphics[width=0.35\linewidth]{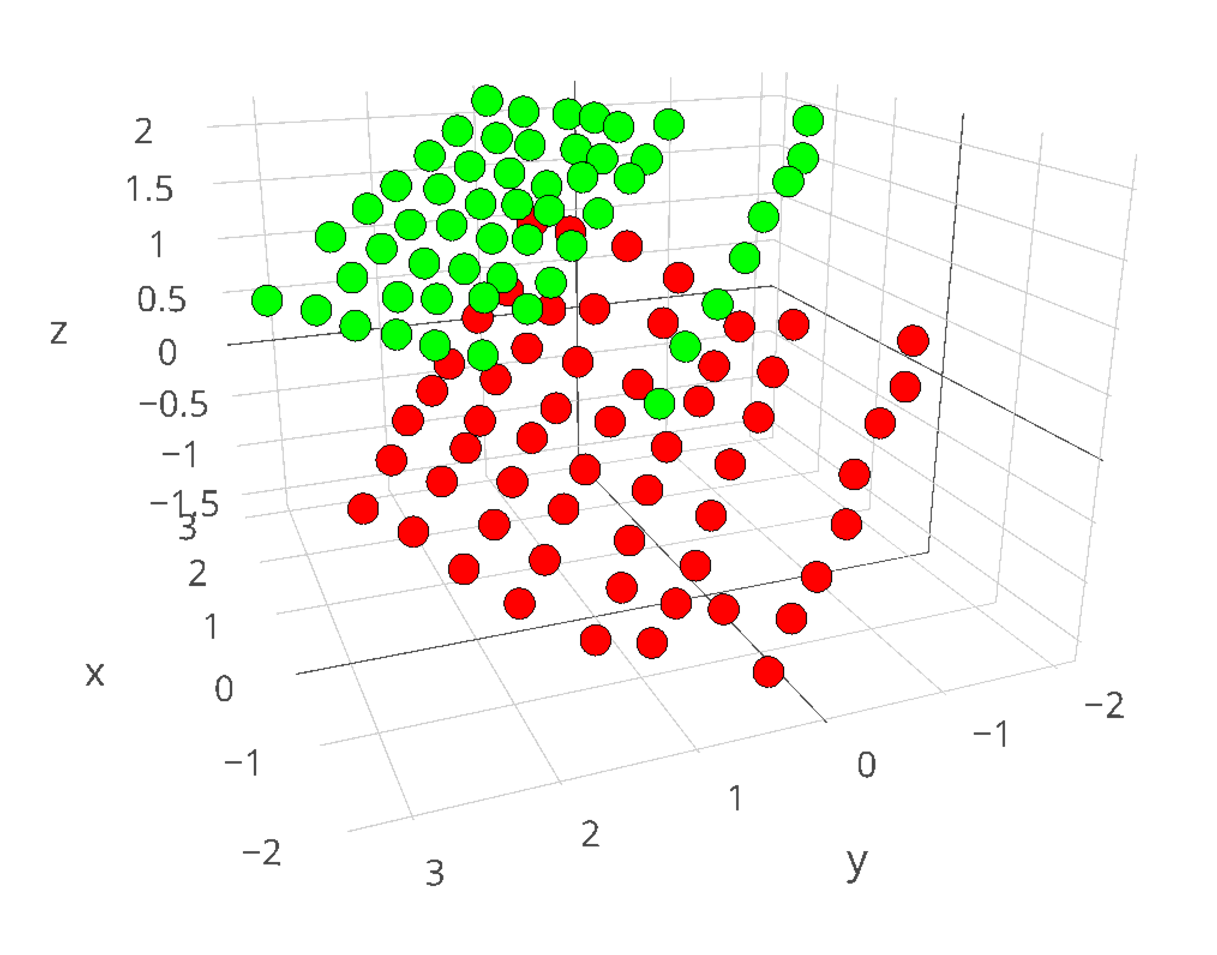}
		 }
		\caption{\textbf{Left:} Plot showing probability of interacting with key and door objects during a self-play episode, as self-play training progresses. Note that the task distribution proposed by Alice changes: initially tasks are only movements (i.e. low chance of either key/door interaction), but chances of collecting the key steeply increases after just a few epochs, and going through the door gradually increases. \textbf{Right:} Goal embeddings learned by Bob for a particular maze. The green (red) points correspond to states where the door locked (unlocked).}
		\label{fig:pick_goals}
\end{figure}

Alice learns to propose increasingly complex tasks throughout self-play. Initially, Alice's tasks are movement based, but by the end of self-play training Alice has explored the space of tasks afforded by the KeyDoor environment and frequently unlocks the door and navigates to the second room. This is illustrated in \fig{pick_goals} (left) which plots the frequency of different types of tasks proposed by Alice. Empirically, we see that the corresponding goal space learned during self-play training reflects the controllable aspects of the environment. \fig{pick_goals} (right) visualizes the learned goal embedding by plotting $\phi(s)$ for each possible state, $s$, of a particular maze instance. Two distinct planes are evident in the goal embedding corresponding to states in which the door is locked (green) and unlocked (red). Within each plane the the spatial structure of the grid world is evident.

We then train Charlie on the KeyDoor task, comparing against REINFORCE \cite{Williams92simplestatistical} and self-play \cite{Sukhbaatar18} algorithms. Our hierarchical self-play model outperforms REINFORCE by a significant margin as shown in \fig{keydoor}(right). Our improvement over the self-play baseline indicates that it is not merely the unsupervised training that provides the boost in performance. Rather, the hierarchy introduced by Charlie is crucial for good performance.

\subsection{Mujoco Ant}
Next, we apply our method to the Ant environment from \cite{duan2016benchmarking}, where the agent controls a four-legged ant-like robot. The agent takes as an observation a 125 dimensional vector that contains location, velocity and joint angles. The action is a 8 dimensional vector that controls the 8 joints of the ant. Although the actions are continuous, we discretize them into 5 bins. 

In self-play, we define the distance function $D$ as a physical distance in the $x-y$ plane, and set $\epsilon=0.25$. This means Bob only needs to reach a location with distance less than $0.25$ from the Alice's last location. Alice always takes $T_A=50$ steps, and Bob is allowed up to $T_B=70$ steps. 
Like the MazeBase experiments, we allow 4 games per episode and use discount $\lambda=0.7$. We set the dimension of goal embedding $K=2$, and use the difference version of the goal embedding function $E$.

\begin{figure}[h!]
	    \centerline{%
		    \includegraphics[width=0.3\linewidth]{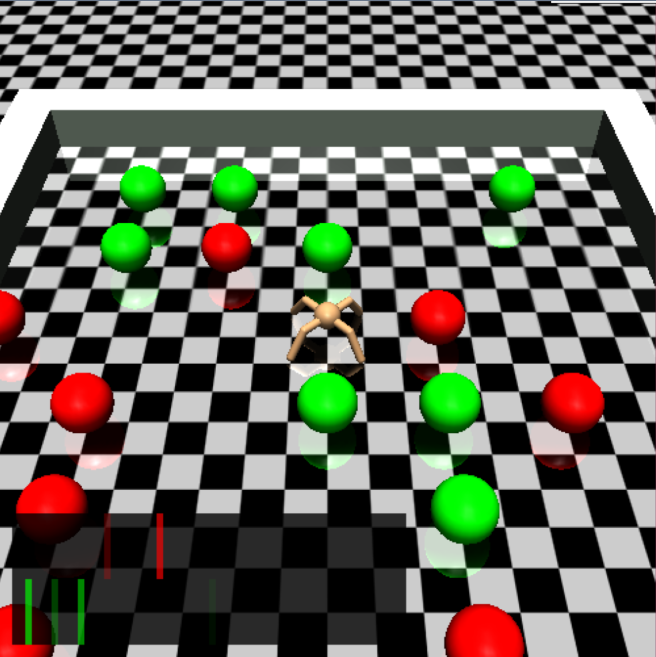}
		    \includegraphics[width=0.65\linewidth]{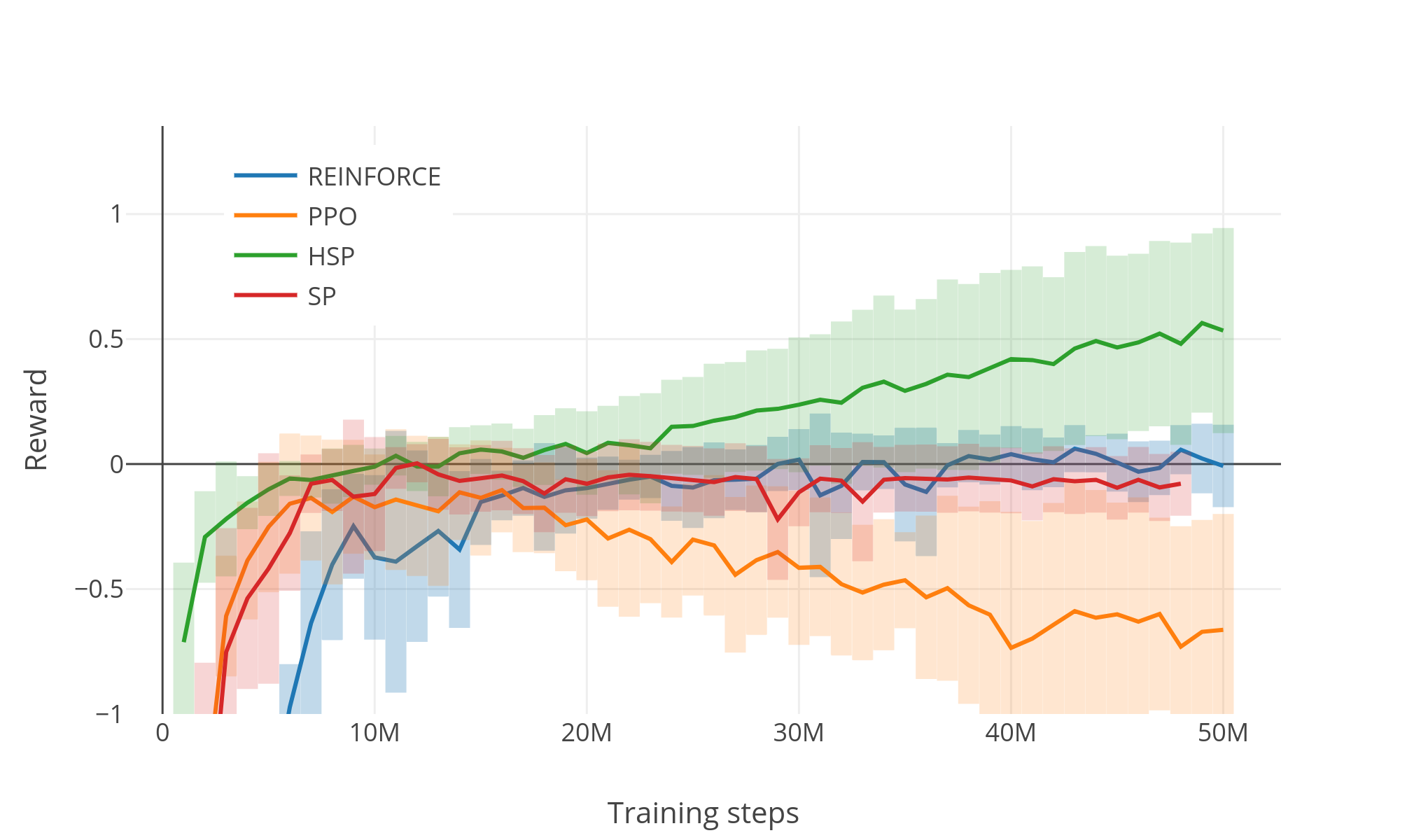}
		 }
		\caption{AntGather task. \textbf{Left:} Example of the environment. \textbf{Right:} Comparison of our approach HSP to other baselines.}
		\label{fig:ant}
\end{figure}

We train Charlie on the more complicated AntGather task \cite{duan2016benchmarking}. As shown in \fig{ant} (left), the ant is placed in a small arena with green and red objects. The object locations are randomized in every episode. When the ant touches an object , the object disappears and the agent gets a reward of $+1$ ($-1$) if the object was green (red). Each episode terminates after 1000 steps. In addition, if the ant jumps too high (in the $z$-axis), it gets penalty of $-10$ and the episode terminates. This makes the task more challenging because it inhibits exploration and introduces a local minima where the ant learns to stay still to avoid this penalty. This penalty is also present in the self-play phase.  

Compared to the Ant environment, the observation of AntGather task includes an additional 20 dimensional vector containing sensory values for detecting nearby objects (see \cite{duan2016benchmarking} for more details). While Charlie takes this full observation as input, we give Bob an observation without those sensory values\footnote{During self-play on the Ant task, the state lacks the sensory input dimensions. Hence Bob would not know what to do with them in the AntGather task.}. We set $T_C=50$, so Charlie's single action corresponds to 50 steps in the environment, and he is allowed to take 20 actions in an episode.

\begin{figure}[h!]
	    \centerline{%
        	\includegraphics[width=0.33\linewidth]{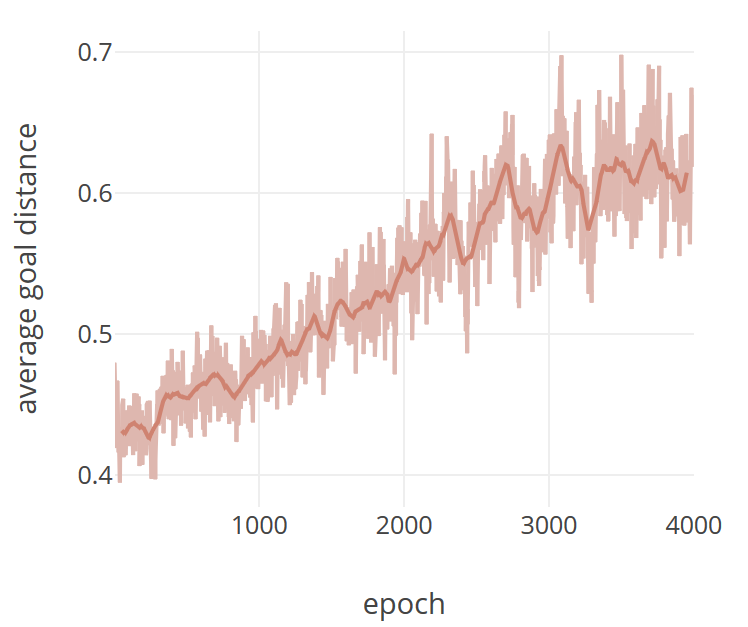}
		    \includegraphics[width=0.3\linewidth]{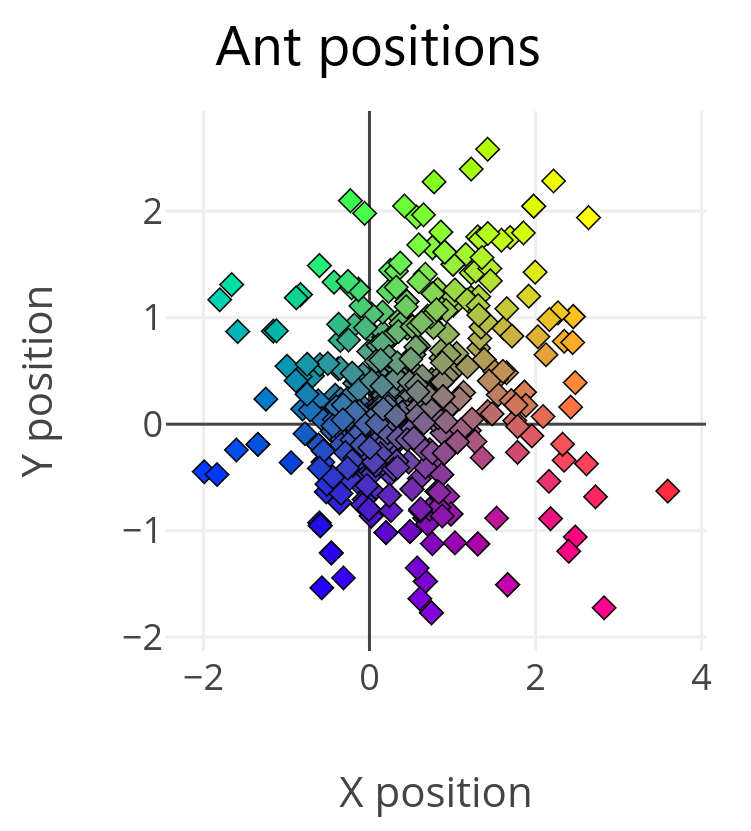}\hspace{2mm}
\includegraphics[width=0.33\linewidth]{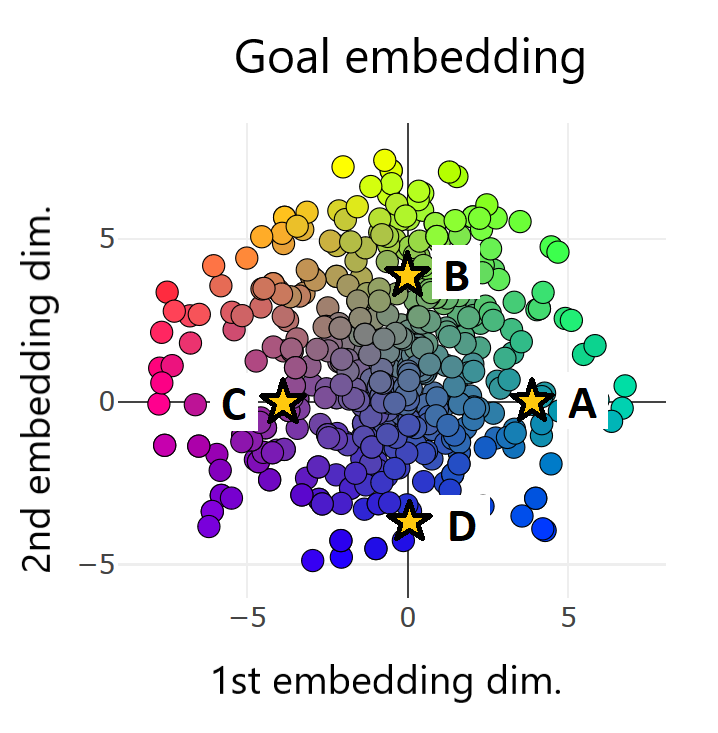}
		 }
		\caption{\textbf{Left:} The average distance of tasks proposed by Alice during training. \textbf{Center:} Goal locations proposed by Alice by the end of training. \textbf{Right}: Corresponding goal embeddings. Their color comes from their location, so we can see the goal space resembles spatial space.}
		\label{fig:ant_goal}
\end{figure}

As a baseline, we compare our model to the REINFORCE \cite{Williams92simplestatistical}, PPO \cite{Schulman2017ProximalPO}, and self-play\footnote{The self-play baseline has almost the same number of unsupervised training steps as our model.} \cite{Sukhbaatar18} algorithms. We use an open-source implementation\footnote{ \url{https://github.com/ikostrikov/pytorch-a2c-ppo-acktr}} of PPO, and run it with the recommended hyper-parameters. As validation, we run it on the Ant task, where it outperforms our REINFORCE baseline by a large margin. However, on the AntGather task, it failed to learn at all. We note that Duan \etal \cite{duan2016benchmarking} also ran several other RL algorithms including TRPO on this task, and all of them failed to learn. The REINFORCE and self-play baselines also fail to obtain a positive reward. In contrast, our model obtains an average reward of 0.5 after training, outperforming the baselines as shown in \fig{ant} (right).

In \fig{ant_goal} (left), we show the average distance of tasks proposed by Alice during self-play training. We can see that Alice and Bob are learning to travel farther as training progresses.
In \fig{ant_goal} (center), we show the actual goal XY locations proposed by Alice by the end of training. It is evident that Alice proposes diverse goal positions in all directions. Large entropy regularization on Alice's policy was important for maintaining this diversity. For every proposed task $s^*$ in \fig{ant_goal} (center), we plot their corresponding goal embedding $\phi(s^*)$ in \fig{ant_goal} (right). The points are colored by their locations from \fig{ant_goal} (center), so we can see the goal embedding captures the spatial structure of the state (up to an arbitrary global transformation) by the end of the training.

\begin{figure}[h!]
	    \centerline{%
		    \includegraphics[width=0.24\linewidth]{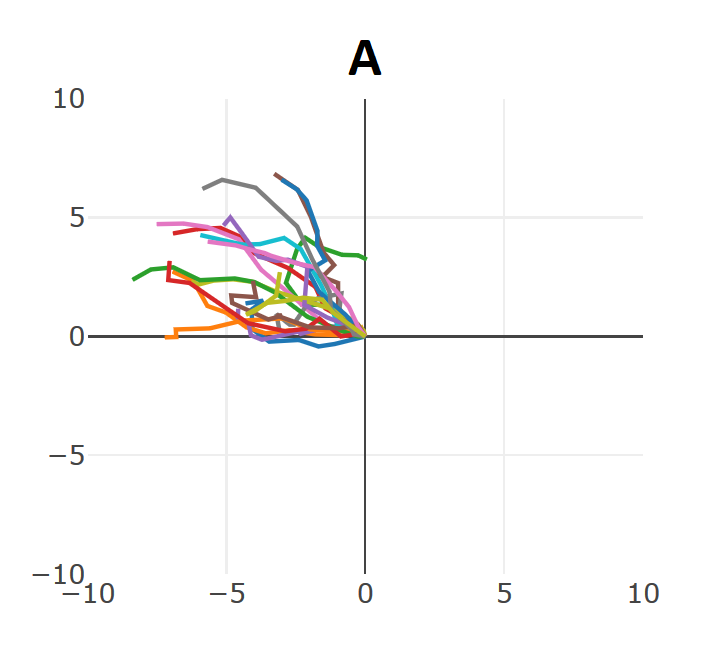}
		    \includegraphics[width=0.24\linewidth]{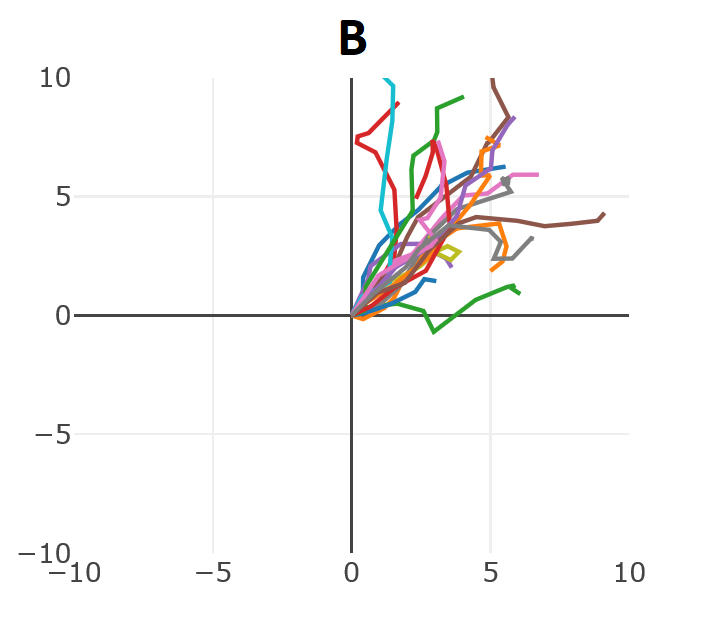}
		    \includegraphics[width=0.24\linewidth]{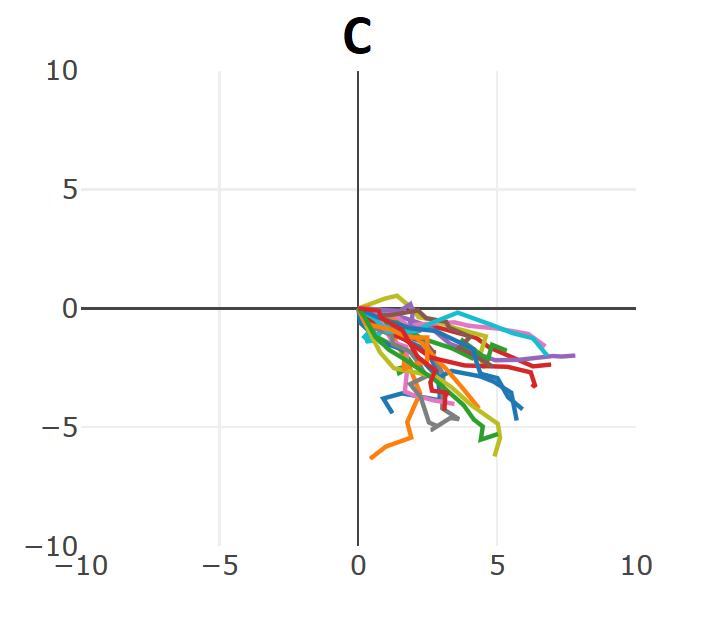}
		    \includegraphics[width=0.24\linewidth]{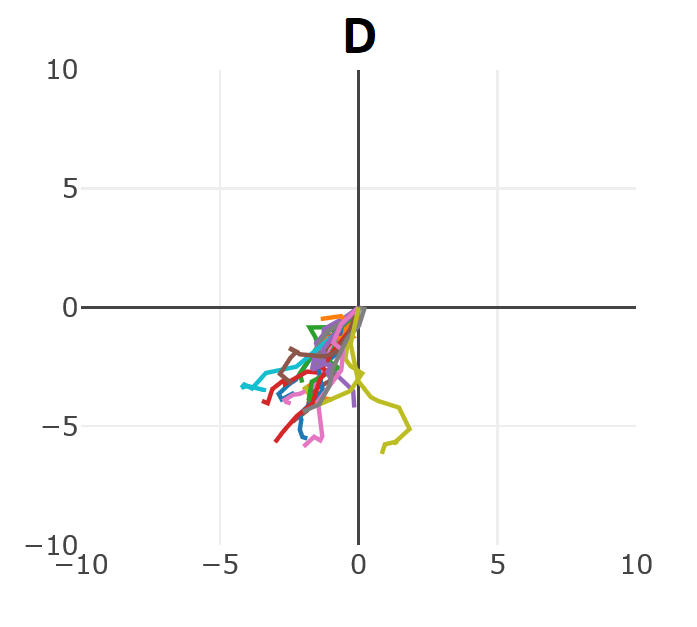}
		 }
		\caption{Trajectories taken by Bob in XY space when given selected goals from \fig{ant_goal}(right) (stars A,B,C,D) as input. Consider, for example, goal A. This corresponds to a target location in the upper left (cyan colored diamonds in \fig{ant_goal} (center)). We see that Bob's trajectories for goal A move in that direction also. Additionally, the total distance covered by Bob is far greater than that traversed during self-play, demonstrating Charlie's ability to reuse Bob's policy and also generalization of the state embedding $\phi(s)$.}

		\label{fig:ant_trace}
\end{figure}

Next, we choose 4 fixed points from \fig{ant_goal}  (right) and fed them as a goal to Bob for 10 times with $T_C=50$. As shown in \fig{ant_trace}, Bob travels in different directions depending on which  goal is chosen. This shows that it is possible to control Bob using the goal vector. There is also diversity inside each goal due to the stochasticity of Bob's policy. Another interesting point is that Bob managed to travel much longer average distances (6 or more) with those fixed goals, even though during self-play, he only traveled distances up to 0.7 and only visited a small spatial region shown in \fig{ant_goal} (center). This indicates that Bob's policy is capable of generalizing to unseen states in some degree.

\section{Discussion}
We have proposed a novel approach for learning goal embeddings that relies on the concept of unsupervised self-play. These can then be utilized in a hierarchical RL framework to speed exploration on complex tasks with sparse reward.   Experiments on AntGather demonstrate the ability of the resulting hierarchical controller to move the Ant long distances to obtain reward, unlike non-hierarchical policy gradient methods.

One limitation of our self-play approach is that the choice of $D$ (the distance function used to decide if the self-play task has been completed successfully or not) requires some domain knowledge. For example, in the Mujoco Ant tasks, we chose $D$ to only care about $x-y$ location, not height and joint angles. 

Although REINFORCE was used in our experiments, more sophisticated policy gradients or Q-learning could be used instead at both the level of the unsupervised self-play pre-training and the test task reinforcement learning.    Another future direction is to expand the self-play concept to the higher-level controller, where a meta-Alice and meta-Bob would play against one another, passing sub-goals to low-level Alice and Bob. Meta-Bob could be used as pre-trained version of Charlie, so reducing the supervision required on the target task. 

\medskip
\bibliography{paper}
\bibliographystyle{ieee}

\end{document}